\documentclass[11pt]{article}
\usepackage{acl2016}
\usepackage{times}
\usepackage{url}
\usepackage{latexsym}
\usepackage{bm}
\usepackage{amssymb,amsmath,amsthm}
\usepackage{amsfonts}
\usepackage{graphicx}
\usepackage{algorithm}
\usepackage{algorithmic}
\usepackage{multirow}
\usepackage[american]{babel}
\usepackage{balance}

\newcommand{\argmax}{\mathop{\rm arg~max}\limits}
\aclfinalcopy 
\def\aclpaperid{320} 

\title{Neural Word Segmentation Learning for Chinese}

\author{Deng Cai \textnormal{and} Hai Zhao\thanks{$\;$Corresponding author. This paper was partially supported by Cai Yuanpei Program (CSC No. 201304490199 and No. 201304490171), National Natural Science Foundation of China (No. 61170114 and No. 61272248), National Basic Research Program of China (No. 2013CB329401), Major Basic Research Program of Shanghai Science and Technology Committee (No. 15JC1400103), Art and Science Interdisciplinary Funds of Shanghai Jiao Tong University (No. 14JCRZ04), and Key Project of National Society Science Foundation of China (No. 15-ZDA041).} \\
	Department of Computer Science and Engineering \\
	Key Lab of Shanghai Education Commission for Intelligent Interaction and Cognitive Engineering
	\\Shanghai Jiao Tong University, Shanghai, China \\
	{\tt thisisjcykcd@gmail.com, zhaohai@cs.sjtu.edu.cn} }
\date{}

\begin{document}
\maketitle
\begin{abstract}
Most previous approaches to Chinese word segmentation formalize this problem as a character-based sequence labeling task where only contextual information within fixed sized local windows and simple interactions between adjacent tags can be captured. In this paper, we propose a novel neural framework which thoroughly eliminates context windows and can utilize complete segmentation history. Our model employs a gated combination neural network over characters to produce distributed representations of word candidates, which are then given to a long short-term memory (LSTM) language scoring model. Experiments on the benchmark datasets show that without the help of feature engineering as most existing approaches, our models achieve competitive or better performances with previous state-of-the-art methods.
\end{abstract}
\section{Introduction}
\begin{table*}[t]
	\centering
	\scalebox{1.}{
	\begin{tabular}{|c|c|c|c|c|}
	\hline
	\multicolumn{2}{|c|}{Models} &Characters&Words&Tags\\
	\cline{1-5}
	\multirow{2}{*}{character based}&\cite{zheng-chen-xu:2013:EMNLP}, \ldots&$c_{i-2},c_{i-1},c_{i},c_{i+1},c_{i+2}$&-&$t_{i-1}t_{i}$\\
	\cline{2-5}
	&\cite{chen-EtAl:2015:EMNLP2}&$c_0,c_1,\ldots, c_i,c_{i+1},c_{i+2}$&-&$t_{i-1}t_{i}$\\
	\hline
	\multirow{2}{*}{word based} &\cite{zhang-clark:2007:ACLMain}, \ldots&$c \text{ in } w_{j-1},w_j,w_{j+1}$&$w_{j-1},w_j,w_{j+1}$&-\\
	\cline{2-5}
	&Ours&$c_0,c_1,\ldots ,c_{i}$&$w_0,w_1,\ldots ,w_{j}$&-\\
	\hline
	\end{tabular}}
	\label{feature-window}
	\caption{Feature windows of different models. $i(j)$ indexes the current character(word) that is under scoring.}
\end{table*}
Most east Asian languages including Chinese are written without explicit word delimiters, therefore, word segmentation is a preliminary step for processing those languages. Since \newcite{xue2003chinese}, most methods formalize the Chinese word segmentation (CWS) as a sequence labeling problem with character position tags, which can be handled with supervised learning methods such as Maximum Entropy \cite{berger1996maximum,low2005maximum} and Conditional Random Fields \cite{lafferty2001conditional,peng2004chinese,zhao2006improved}. However, those methods heavily depend on the choice of handcrafted features.

Recently, neural models have been widely used for NLP tasks for their ability to minimize the effort in feature engineering. For the task of CWS, \newcite{zheng-chen-xu:2013:EMNLP} adapted the general neural network architecture for sequence labeling proposed in \cite{collobert2011natural}, and used character embeddings as input to a two-layer network. \newcite{pei-ge-chang:2014:P14-1} improved upon \cite{zheng-chen-xu:2013:EMNLP} by explicitly modeling the interactions between local context and previous tag. \newcite{chen-EtAl:2015:ACL-IJCNLP5} proposed a gated recursive neural network to model the feature combinations of context characters. \newcite{chen-EtAl:2015:EMNLP2} used an LSTM architecture to capture potential long-distance dependencies, which alleviates the limitation of the size of context window but introduced another window for hidden states.
	
Despite the differences, all these models are designed to solve CWS by assigning labels to the characters in the sequence one by one. At each time step of inference, these models compute the tag scores of character based on (i) context features within a fixed sized local window and (ii) tagging history of previous \textit{one}. 
	
Nevertheless, the tag-tag transition is insufficient to model the complicated influence from previous segmentation decisions, though it could sometimes be a crucial clue to later segmentation decisions. The fixed context window size, which is broadly adopted by these methods for feature engineering, also restricts the flexibility of modeling diverse distances. Moreover, word-level information, which is being the greater granularity unit as suggested in \cite{huang2006essential}, remains unemployed.

To alleviate the drawbacks inside previous methods and release those inconvenient constrains such as the fixed sized context window, this paper makes a latest attempt to re-formalize CWS as a direct segmentation learning task. Our method does not make tagging decisions on individual characters, but directly evaluates the relative likelihood of different segmented sentences and then search for a segmentation with the highest score. To feature a segmented sentence, a series of distributed vector representations \cite{bengio2003neural} are generated to characterize the corresponding word candidates. Such a representation setting makes the decoding quite different from previous methods and indeed much more challenging, however, more discriminative features can be captured. 

Though the vector building is word centered, our proposed scoring model covers all three processing levels from character, word until sentence. First, the distributed representation starts from character embedding, as in the context of word segmentation, the $n$-gram data sparsity issue makes it impractical to use word vectors immediately. Second, as the word candidate representation is derived from its characters, the inside character structure will also be encoded, thus it can be used to determine the word likelihood of its own. Third, to evaluate how a segmented sentence makes sense through word interacting, an LSTM \cite{hochreiter1997long} is used to chain together word candidates incrementally and construct the representation of partially segmented sentence at each decoding step, so that the coherence between next word candidate and previous segmentation history can be depicted. 

To our best knowledge, our proposed approach to CWS is the first attempt which explicitly models the entire contents of the segmenter's state, including the complete history of both segmentation decisions and input characters. The comparisons of feature windows used in different models are shown in Table \ref{feature-window}. Compared to both sequence labeling schemes and word-based models in the past, our model thoroughly eliminates context windows and can capture the complete history of segmentation decisions, which offers more possibilities to effectively and accurately model segmentation context.
\section{Overview}
\begin{figure}[t]
	\centering
	\includegraphics[scale=0.40]{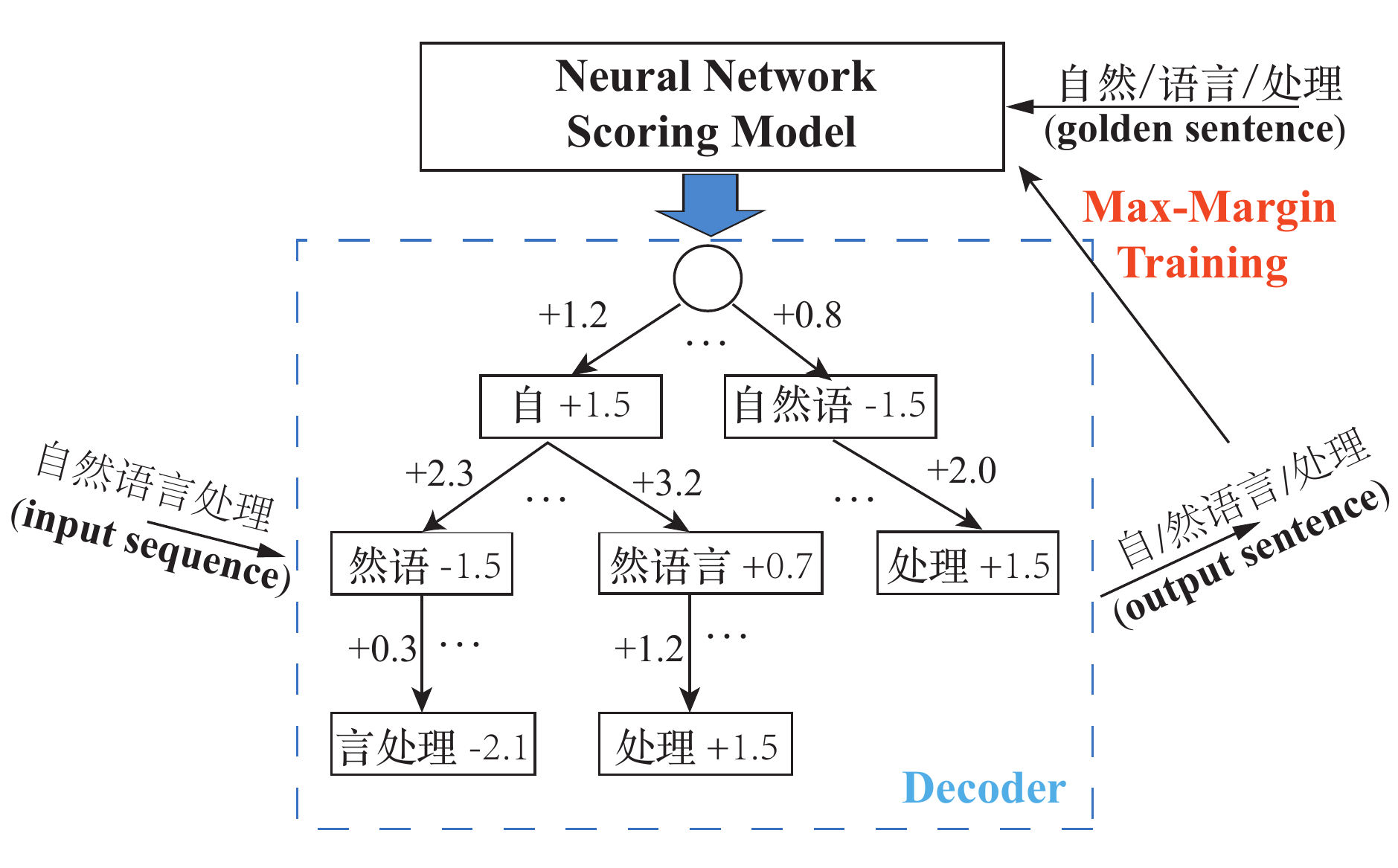}
	\caption{Our framework.}
\end{figure}
We formulate the CWS problem as finding a mapping from an input character sequence $x$ to a word sequence $y$, and the output sentence $y^*$ satisfies:
\begin{equation}
\nonumber
y^* = \argmax_{y\in \text{GEN}(x)}(\sum_{i=1}^{n}\text{score}(y_i|y_1,\cdots,y_{i-1}))
\end{equation}
\begin{figure*}[t]
	\centering
	\label{f3}
	\scalebox{0.8}[0.8]{
	\includegraphics{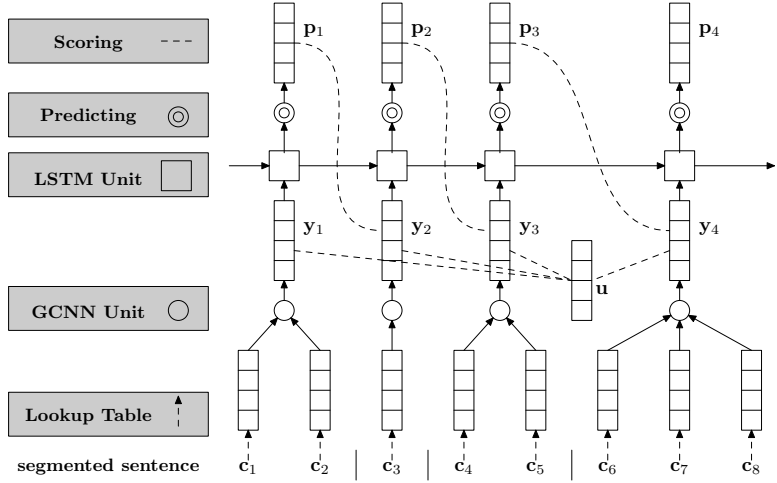}}
	\caption{Architecture of our proposed neural network scoring model, where $\mathbf{c}_i$ denotes the $i$-th input character, $\mathbf{y}_j$ denotes the learned representation of the $j$-th word candidate, $\mathbf{p}_k$ denotes the prediction for the $(k+1)$-th word candidate and $\mathbf{u}$ is the trainable parameter vector for scoring the likelihood of individual word candidates.}
\end{figure*}
where $n$ is the number of word candidates in $y$, and GEN($x$) denotes the set of possible segmentations for an input sequence $x$. Unlike all previous works, our scoring function is sensitive to the complete contents of partially segmented sentence.

As shown in Figure 1, to solve CWS in this way, a neural network scoring model is designed to evaluate the likelihood of a segmented sentence. Based on the proposed model, a decoder is developed to find the segmented sentence with the highest score. Meanwhile, a max-margin method is utilized to perform the training by comparing the structured difference of decoder output and the golden segmentation. The following sections will introduce each of these components in detail.
\section{Neural Network Scoring Model}
The score for a segmented sentence is computed by first mapping it into a sequence of word candidate vectors, then the scoring model takes the vector sequence as input, scoring on each word candidate from two perspectives: (1) how likely the word candidate itself can be recognized as a legal word; (2) how reasonable the link is for the word candidate to follow previous segmentation history immediately. After that, the word candidate is appended to the segmentation history, updating the state of the scoring system for subsequent judgements. Figure 2 illustrates the entire scoring neural network.
\subsection{Word Score}
\paragraph{Character Embedding.} While the scores are decided at the word-level, using word embedding \cite{bengio2003neural,plwang:2016} immediately will lead to a remarkable issue that rare words and out-of-vocabulary words will be poorly estimated \cite{kim2015character}. In addition, the character-level information inside an $n$-gram can be helpful to judge whether it is a true word. Therefore, a lookup table of character embeddings is used as the bottom layer. 

Formally, we have a character dictionary $\mathit{D}$ of size $|\mathit{D}|$. Then each character $\mathit{c}\in \mathit{D}$ is represented as a real-valued vector (character embedding) $\mathbf{c}\in \mathbb{R}^d$, where $d$ is the dimensionality of the vector space. The character embeddings are then stacked into an embedding matrix $\mathbf{M}\in \mathbb{R}^{d\times |D|}$. For a character $\mathit{c}\in \mathit{D}$, its character embedding $\mathbf{c}\in \mathbb{R}^d$ is retrieved by the embedding layer according to its index.

\paragraph{Gated Combination Neural Network.}
In order to obtain word representation through its characters, in the simplest strategy, character vectors are integrated into their word representation using a weight matrix $\mathbf{W}^{(L)}$ that is shared across all words with the same length $L$, followed by a non-linear function $g(\cdot)$. Specifically, $\mathbf{c}_i$ $(1\le i\le L)$ are $d$-dimensional character vector representations respectively, the corresponding word vector $\mathbf{w}$ will be $d$-dimensional as well:
\begin{equation}
	\mathbf{w}=g( \mathbf{W}^{(L)} \left[\begin{array}{c}
	\mathbf{c}_1\\
	\vdots\\
	\mathbf{c}_{L}
	\end{array}\right])
\end{equation}
where $\mathbf{W}^{(L)}\in \mathbb{R}^{d\times Ld}$ and $g$ is a non-linear function as mentioned above.
	
Although the mechanism above seems to work well, it can not sufficiently model the complicated combination features in practice, yet.
	
Gated structure in neural network can be useful for hybrid feature extraction according to \cite{chen-EtAl:2015:ACL-IJCNLP5,chung2014empirical,cho-EtAl:2014:EMNLP2014}, we therefore propose a gated combination neural network (GCNN) especially for character compositionality which contains two types of gates, namely \textit{reset gate} and \textit{update gate}. Intuitively, the reset gates decide which part of the character vectors should be mixed while the update gates decide what to preserve when combining the characters information.
\begin{figure}[t]
	\centering
	\label{f4}
	\includegraphics{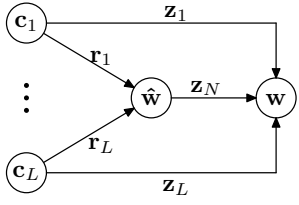}
	\caption{Gated combination neural network.}
\end{figure}	
Concretely, for words with length $L$, the word vector $\mathbf{w}\in \mathbb{R}^{d}$ is computed as follows:
\begin{equation}
		\nonumber
		\mathbf{w}=\mathbf{z}_N\odot \mathbf{\hat{w}}+\sum_{i=1}^{L}\mathbf{z}_{i}\odot \mathbf{c}_i
\end{equation}
where $\mathbf{z}_N,\mathbf{z}_i$ $(1\le i\le L)$ are update gates for new activation $\mathbf{\hat{w}}$ and governed characters respectively, and $\odot$ indicates element-wise multiplication.
	
The new activation $\mathbf{\hat{w}}$ is computed as:
\begin{equation}
	\nonumber
	\mathbf{\hat{w}}=\tanh(\mathbf{W}^{(L)}\left[\begin{array}{c}
	\mathbf{r}_1\odot \mathbf{c}_1\\
	\vdots\\
	\mathbf{r}_{L}\odot \mathbf{c}_{L}\\
	\end{array}\right])
\end{equation}
where $\mathbf{W}^{(L)}\in \mathbb{R}^{d\times Ld}$ and $\mathbf{r}_i\in \mathbb{R}^d$ $(1\le i\le L)$ are the reset gates for governed characters respectively, which can be formalized as:
\begin{equation}
	\nonumber
	\left[\begin{array}{c}
		\mathbf{r}_1\\
		\vdots\\
		\mathbf{r}_{L}\\
	\end{array}\right]=\sigma(\mathbf{R}^{(L)}\left[\begin{array}{c}
	\mathbf{c}_1\\
	\vdots\\
	\mathbf{c}_{L}
	\end{array}\right])
\end{equation}
where $\mathbf{R}^{(L)}\in \mathbb{R}^{Ld\times Ld}$ is the coefficient matrix of reset gates and $
\sigma$ denotes the sigmoid function.

The update gates can be formalized as:
\begin{equation}
	\nonumber
	\left[\begin{array}{c}
		\mathbf{z}_N\\
		\mathbf{z}_1\\
		\vdots\\
		\mathbf{z}_{L}\\
	\end{array}\right]=\exp(\mathbf{U}^{(L)}\left[\begin{array}{c}
	\mathbf{\hat{w}}\\
	\mathbf{c}_1\\
	\vdots\\
	\mathbf{c}_{L}
\end{array}\right])\odot\left[\begin{array}{c}
1/\mathbf{Z}\\
1/\mathbf{Z}\\
\vdots\\
1/\mathbf{Z}
\end{array}\right]
\end{equation}
where $\mathbf{U}^{(L)}\in \mathbb{R}^{(L+1)d\times (L+1)d}$ is the coefficient matrix of update gates, and $\mathbf{Z}\in \mathbb{R}^{d}$ is the normalization vector,
\begin{equation}
	\nonumber
	\mathbf{Z}_k=\sum_{i=1}^{L}[\exp(\mathbf{U}^{(L)}\left[\begin{array}{c}
		\mathbf{\hat{w}}\\
		\mathbf{c}_1\\
		\vdots\\
		\mathbf{c}_{L}
	\end{array}\right] )]_{d\times i+k}
\end{equation}
where $0\le k< d$.

According to the normalization condition, the update gates are constrained by:
\begin{equation}
	\nonumber
	\mathbf{z}_{N}+\sum_{i=1}^{L}\mathbf{z}_{i} = \mathbf{1}
\end{equation}

The gated mechanism is capable of capturing both character and character interaction characteristics to give an efficient word representation (See Section \ref{MA}).

\paragraph{Word Score.} Denote the learned vector representations for a segmented sentence $y$ with $[\mathbf{y}_1,\mathbf{y}_2,\cdots,\mathbf{y}_n]$, where $n$ is the number of word candidates in the sentence. \textit{word score} will be computed by the dot products of vector $\mathbf{y}_i (1\le i\le n)$ and a trainable parameter vector $\mathbf{u}\in \mathbb{R}^d$.
\begin{equation}
\mathrm{Word\_Score}(\mathbf{y}_i) = \mathbf{u}\cdot\mathbf{y}_i
\end{equation}
It indicates how likely a word candidate by itself is to be a true word.
\subsection{Link Score}
Inspired by the recurrent neural network language model (RNN-LM) \cite{mikolov2010recurrent,sundermeyer2012lstm}, we utilize an LSTM system to capture the coherence in a segmented sentence.

\paragraph{Long Short-Term Memory Networks.} The LSTM neural network \cite{hochreiter1997long} is an extension of the recurrent neural network (RNN), which is an effective tool for sequence modeling tasks using its hidden states for history information preservation. At each time step $t$, an RNN takes the input $\mathbf{x}_t$ and updates its recurrent hidden state $\mathbf{h}_t$ by
\begin{equation}
\nonumber
\mathbf{h}_t=g(\mathbf{U}\mathbf{h}_{t-1}+\mathbf{W}\mathbf{x}_t+\mathbf{b})
\end{equation}
where $g$ is a non-linear function.

Although RNN is capable, in principle, to process arbitrary-length sequences, it can be difficult to train an RNN to learn long-range dependencies due to the vanishing gradients. LSTM addresses this problem by introducing a memory cell to preserve states over long periods of time, and controls the update of hidden state and memory cell by three types of gates, namely \textit{input gate}, \textit{forget gate} and \textit{output gate}. Concretely, each step of LSTM takes input $\mathbf{x}_t,\mathbf{h}_{t-1},\mathbf{c}_{t-1}$ and produces $\mathbf{h}_t,\mathbf{c}_t$ via the following calculations:
\begin{equation}
\nonumber
\begin{split}
&\mathbf{i}_t=\sigma(\mathbf{W}^i\mathbf{x}_t+\mathbf{U}^i\mathbf{h}_{t-1}+\mathbf{b}^i) \\
&\mathbf{f}_t=\sigma(\mathbf{W}^f\mathbf{x}_t+\mathbf{U}^f\mathbf{h}_{t-1}+\mathbf{b}^f) \\
&\mathbf{o}_t=\sigma(\mathbf{W}^o\mathbf{x}_t+\mathbf{U}^o\mathbf{h}_{t-1}+\mathbf{b}^o) \\
&\mathbf{\hat{c}}_t=\tanh(\mathbf{W}^c\mathbf{x}_t+\mathbf{U}^c\mathbf{h}_{t-1}+\mathbf{b}^c) \\
&\mathbf{c}_t = \mathbf{f}_t\odot \mathbf{c}_{t-1}+\mathbf{i}_t\odot \mathbf{\hat{c}}_t\\
&\mathbf{h}_t=\mathbf{o}_t\odot \tanh(\mathbf{c}_t)\\
\end{split}
\end{equation}
where $\sigma,\odot$ are respectively the element-wise sigmoid function and multiplication, $\mathbf{i}_t,\mathbf{f}_t,\mathbf{o}_t,\mathbf{c}_t$ are respectively the input gate, forget gate, output gate and memory cell activation vector at time $t$, all of which have the same size as hidden state vector $\mathbf{h}_t\in \mathbb{R}^H$.
\begin{figure}[t]
	\centering
	\label{f6}
	\includegraphics[scale=0.8]{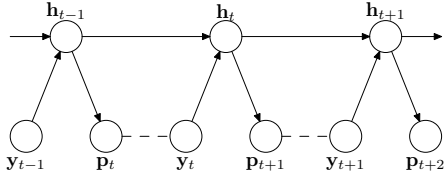}
	\caption{Link scores (dashed lines).}
\end{figure}

\paragraph{Link Score.} LSTMs have been shown to outperform RNNs on many NLP tasks, notably language modeling \cite{sundermeyer2012lstm}.

In our model, LSTM is utilized to chain together word candidates in a left-to-right, incremental manner. At time step $t$, a prediction $\mathbf{p}_{t+1}\in \mathbb{R}^d$ about next word $\mathbf{y}_{t+1}$ is made based on the hidden state $\mathbf{h}_t$:
\begin{equation}
	\nonumber
	\mathbf{p}_{t+1} = \tanh(\mathbf{W}^p\mathbf{h}_t+\mathbf{b}^p)
\end{equation}
\textit{link score} for next word $\mathbf{y}_{t+1}$ is then computed as:
\begin{equation}
\mathrm{Link\_Score}(\mathbf{y}_{t+1}) = \mathbf{p}_{t+1}\cdot\mathbf{y}_{t+1}
\end{equation}

Due to the structure of LSTM, the prediction vector $\mathbf{p}_{t+1}$ carries useful information detected from the entire segmentation history, including previous segmentation decisions. In this way, our model gains the ability of sequence-level discrimination rather than local optimization.

\subsection{Sentence score} \textit{Sentence score} for a segmented sentence $y$ with $n$ word candidates is computed by summing up word scores (2) and link scores (3) as follow:
\begin{equation}\label{sentence_score}
s(y_{[1:n]},\theta) = \sum_{t=1}^{n}(\mathbf{u}\cdot\mathbf{y}_t+{\mathbf{p}_t}\cdot\mathbf{y}_t)
\end{equation}
where $\theta$ is the parameter set used in our model.
\section{Decoding}
The total number of possible segmented sentences grows exponentially with the length of character sequence, which makes it impractical to compute the scores of every possible segmentation. In order to get exact inference, most sequence-labeling systems address this problem with a Viterbi search, which takes the advantage of their hypothesis that the tag interactions only exist within adjacent characters (Markov assumption). However, since our model is intended to capture complete history of segmentation decisions, such dynamic programming algorithms can not be adopted in this situation. 
\begin{algorithm}[h]
	\caption{Beam Search.}
	\begin{algorithmic}[1]
		\REQUIRE $\;$model parameters $\theta$ \\ \quad\quad beam size $k$\\ \quad\quad maximum word length $w$ \\ \quad\quad input character sequence $c[1:n]$
		\ENSURE Approx. $k$ best segmentations
		\STATE $\pi[0]$ $\gets$ $\{(score=0,\mathbf{h}=\mathbf{h}_0,\mathbf{c}=\mathbf{c}_0)\}$
		\FOR {$i=1$ to $n$}
		\STATE $\triangleright$ Generate Candidate Word Vectors
		\STATE $X \gets\emptyset$
		\FOR {$j=\max(1,i-w)$ to $i$}
		\STATE $\mathbf{w} = \text{GCNN-Procedure} (c[j:i])$ 
		\STATE $X.\text{add}((index=j-1,word=\mathbf{w}))$
		\ENDFOR
		\STATE $\triangleright$ Join Segmentation
		\STATE $Y$ $\gets$ $\{ $ $y.\text{append}(x) $ $|$ $y\in\pi[x.index]$ and $x\in X\}$
		\STATE $\triangleright$ Filter $k$-Max
		\STATE $\pi[i]$ $\gets$ $k\text{-}\argmax_{y\in Y}y.score$
		\ENDFOR
		\RETURN $\pi[n]$
	\end{algorithmic}
\end{algorithm}

To make our model efficient in practical use, we propose a beam-search algorithm with dynamic programming motivations as shown in Algorithm 1. The main idea is that any segmentation of the first $i$ characters can be separated as two parts, the first part consists of characters with indexes from $0$ to $j$ that is denoted as $y$, the rest part is the word composed by $c[j+1:i]$. The influence from previous segmentation $y$ can be represented as a triple ($y.score$, $y.\mathbf{h}$, $y.\mathbf{c}$), where $y.score$, $y.\mathbf{h}$, $y.\mathbf{c}$ indicate the current score, current hidden state vector and current memory cell vector respectively. Beam search ensures that the total time for segmenting a sentence of $n$ characters is $w\times k\times n$, where $w,k$ are maximum word length and beam size respectively.

\section{Training}
We use the max-margin criterion \cite{taskar2005learning} to train our model. As reported in \cite{kummerfeld-bergkirkpatrick-klein:2015:EMNLP}, the margin methods generally outperform both likelihood and perception methods. For a given character sequence $x^{(i)}$, denote the correct segmented sentence for $x^{(i)}$ as $y^{(i)}$. We define a structured margin loss $\Delta (y^{(i)},\hat{y})$ for predicting a segmented sentence $\hat{y}$:
\begin{equation}
	\nonumber
	\Delta (y^{(i)},\hat{y})=\sum_{t=1}^{m}\mu\mathbf{1}\{y^{(i),t}\neq \hat{y}^t\}
\end{equation}
where $m$ is the length of sequence $x^{(i)}$ and $\mu$ is the discount parameter. The calculation of margin loss could be regarded as to count the number of incorrectly segmented characters and then multiple it with a fixed discount parameter for smoothing. Therefore, the loss is proportional to the number of incorrectly segmented characters.

Given a set of training set $\varOmega$, the regularized objective function is the loss function $J(\theta)$ including an $\ell_2$ norm term: 
\begin{equation}
	\nonumber
	J(\theta) = \frac{1}{|\varOmega|} \sum_{(x^{(i)},y^{(i)})\in \varOmega} l_i(\theta)+\frac{\lambda}{2}{||\theta||}_2^2
\end{equation}
\begin{equation}
	\nonumber
	l_i(\theta)=\max_{\hat{y}\in \text{GEN}(x^{(i)})}{(s(\hat{y},\theta)+\Delta(y^{(i)},\hat{y})-s(y^{(i)},\theta))}
\end{equation}
where the function $s(\cdot)$ is the sentence score defined in equation (\ref{sentence_score}).

Due to the hinge loss, the objective function is not differentiable, we use a subgradient method \cite{ratliff2007approximate} which computes a gradient-like direction. Following \cite{socher-EtAl:2013:ACL2013}, we use the diagonal variant of AdaGrad \cite{duchi2011adaptive} with minibatchs to minimize the objective. The update for the $i$-th parameter at time step $t$ is as follows:
\begin{equation}
	\nonumber
	\theta_{t,i}=\theta_{t-1,i}-\frac{\alpha}{\sqrt{\sum_{\tau=1}^{t}g_{\tau,i}^2}}g_{t,i}
\end{equation}
where $\alpha$ is the initial learning rate and $g_{\tau,i}\in \mathbb{R}^{|\theta_i|}$ is the subgradient at time step $\tau$ for parameter $\theta_i$.

\section{Experiments}
\subsection{Datasets}
To evaluate the proposed segmenter, we use two popular datasets, PKU and MSR, from the second International Chinese Word Segmentation Bakeoff \cite{emerson2005second}. These datasets are commonly used by previous state-of-the-art models and neural network models.

Both datasets are preprocessed by replacing the continuous English characters and digits with a unique token. All experiments are conducted with standard Bakeoff scoring program\footnote[1]{http://www.sighan.org/bakeoff2003/score} calculating precision, recall, and $\text{F}_1$-score.
\subsection{Hyper-parameters}
\begin{table}[t]
	\centering
	\begin{tabular}{|l|l|}
		\hline
		Character embedding size & $d=50$\\
		Hidden unit number & $H=50$ \\ 
		Initial learning rate & $\alpha=0.2$\\
		Margin loss discount & $\mu=0.2$\\
		Regularization & $\lambda=10^{-6}$\\
		Dropout rate on input layer& $p=0.2$\\
		Maximum word length & $w=4$\\
		\hline
	\end{tabular}
	\caption{Hyper-parameter settings.}
\end{table}
Hyper-parameters of neural network model significantly impact on its performance. To determine a set of suitable hyper-parameters, we divide the training data into two sets, the first 90\% sentences as training set and the rest 10\% sentences as development set. We choose the hyper-parameters as shown in Table 2.

We found that the character embedding size has a limited impact on the performance as long as it is large enough. The size 50 is chosen as a good trade-off between speed and performance. The number of hidden units is set to be the same as the character embedding. Maximum word length determines the number of parameters in GCNN part and the time consuming of beam search, since the words with a length $l>4$ are relatively rare, 0.29\% in PKU training data and 1.25\% in MSR training data, we set the maximum word length to 4 in our experiments.\footnote[2]{This 4-character limitation is just for consistence for both datasets. We are aware that it is a too strict setting, especially which makes additional performance loss in a dataset with larger average word length, i.e., MSR.}

Dropout is a popular technique for improving the performance of neural networks by reducing overfitting \cite{srivastava2014dropout}. We also drop the input layer of our model with dropout rate 20\% to avoid overfitting.
\subsection{Model Analysis}\label{MA}
\begin{figure}[t]
	\centering
	\label{f7}
	\includegraphics{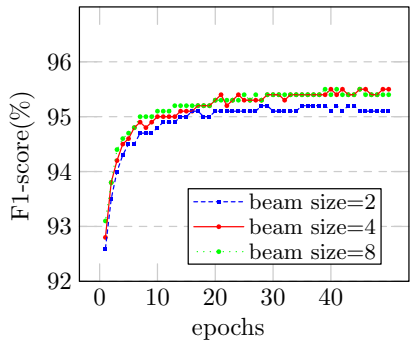}
	\caption{Performances of different beam sizes on PKU dataset.}
\end{figure}
\begin{figure}[t]
	\centering
	\label{f8}
	\includegraphics{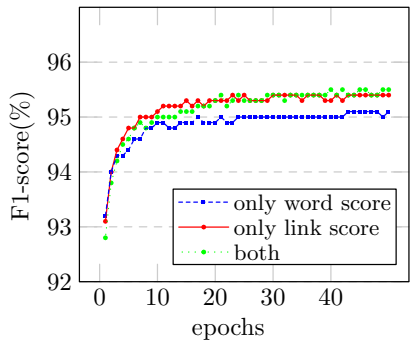}
	\caption{Performances of different score strategies on PKU dataset.}
\end{figure}
\begin{table}[t]
	\centering
	\begin{tabular}{|l|c|c|c|}
		\hline
		models &P&R&F \\
		\hline
		Single layer $(d=50)$& 94.3 &93.7 &94.0\\
		GCNN $(d=50)$ & 95.8 &95.2 &95.5\\
		Single layer $(d=100)$& 94.9 &94.4 &94.7\\
		\hline
	\end{tabular}
	\caption{Performances of different models on PKU dataset.}
\end{table}
\paragraph{Beam Size.} We first investigated the impact of beam size over segmentation performance. Figure 5 shows that a segmenter with beam size 4 is enough to get the best performance, which makes our model find a good balance between accuracy and efficiency.

\paragraph{GCNN.} We then studied the role of GCNN in our model. To reveal the impact of GCNN, we re-implemented a simplified version of our model, which replaces the GCNN part with a single non-linear layer as in equation (1). The results are listed in Table 3, which demonstrate that the performance is significantly boosted by exploiting the GCNN architecture (94.0\% to 95.5\% on $\text{F}_1$-score), while the best performance that the simplified version can achieve is 94.7\%, but using a much larger character embedding size.
\paragraph{Link Score \& Word Score.} We conducted several experiments to investigate the individual effect of link score and word score, since these two types of scores are intended to estimate the sentence likelihood from two different perspectives: the semantic coherence between words and the existence of individual words. The learning curves of models with different scoring strategies are shown in Figure 6. 

The model with only word score can be regarded as the situation that the segmentation decisions are made only based on local window information. The comparisons show that such a model gives moderate performance. By contrast, the model with only link score gives a much better performance close to the joint model, which demonstrates that the complete segmentation history, which can not be effectively modeled in previous schemes, possesses huge appliance value for word segmentation.
\begin{table}[t]
	\centering
	\scalebox{0.9}{
		\begin{tabular}{|c|c|c|c|c|}
			\hline
			&\multicolumn{2}{c|}{PKU} &\multicolumn{2}{c|}{MSR}\\
			\hline
			+Dictionary&ours&theirs&ours&theirs\\
			\hline
			\cite{chen-EtAl:2015:ACL-IJCNLP5} &94.9&\textbf{95.9}&95.8&96.2\\
			\cite{chen-EtAl:2015:EMNLP2} & 94.6&95.7&95.7&\textbf{96.4}\\
			This work &\textbf{95.7}&-&\textbf{96.4}&-\\
			\hline
		\end{tabular}}
		\caption{Comparison of using different Chinese idiom dictionaries.\footnotemark[3]}
\end{table}
\begin{table*}[t]
	\centering
	\begin{tabular}{|c|c|c|c|c|c|c|}
		\hline
		\multirow{2}{*}{Models} & \multicolumn{3}{c|}{PKU} &\multicolumn{3}{c|}{MSR}\\
		\cline{2-7}
		&P&R&F&P&R&F\\
		\hline
		\cite{zheng-chen-xu:2013:EMNLP}&92.8&92.0&92.4&92.9&93.6&93.3\\
		\cite{pei-ge-chang:2014:P14-1}&93.7&93.4&93.5&94.6&94.2&94.4\\
		\cite{chen-EtAl:2015:ACL-IJCNLP5}* &94.6&94.2&94.4&94.6&95.6&95.1\\
		\cite{chen-EtAl:2015:EMNLP2} *&94.6&94.0&94.3&94.5&95.5&95.0\\
		This work&\textbf{95.5}&\textbf{94.9}&\textbf{95.2}&\textbf{96.1}&\textbf{96.7}&\textbf{96.4}\\
		\hline\hline
		\textbf{+Pre-trained character embedding}&&&&&&\\
		\cite{zheng-chen-xu:2013:EMNLP}&93.5&92.2&92.8&94.2&93.7&93.9\\
		\cite{pei-ge-chang:2014:P14-1}&94.4&93.6&94.0&95.2&94.6&94.9\\
		\cite{chen-EtAl:2015:ACL-IJCNLP5}*&94.8&94.1&94.5&94.9&95.9&95.4\\
		\cite{chen-EtAl:2015:EMNLP2}*&95.1&94.4&94.8&95.1&96.2&95.6\\
		This work&\textbf{95.8}&\textbf{95.2}&\textbf{95.5}&\textbf{96.3}&\textbf{96.8}&\textbf{96.5}\\
		\hline	
	\end{tabular}
	\caption{Comparison with previous neural network models. Results with * are from our runs on their released implementations.\footnotemark[5]}
\end{table*}
\begin{table*}[t]
	\centering
	\begin{tabular}{|c|c|c|c|c|}
		\hline
		Models&PKU&MSR&PKU&MSR\\
		\hline
		\cite{tseng2005conditional} & 95.0&96.4&-&-\\
		\cite{zhang-clark:2007:ACLMain} &94.5&97.2&-&-\\
		\cite{zhao2008unsupervised} & 95.4& 97.6&-&-\\
		\cite{sun2009discriminative}& 95.2&97.3&-&-\\
		\cite{sun2012fast}& 95.4&97.4&-&-\\
		\cite{zhang-EtAl:2013:EMNLP1}
		&-&-& 96.1*&97.4*\\
		\hline
		\cite{chen-EtAl:2015:ACL-IJCNLP5} &94.5&95.4& 96.4*&97.6*\\
		\cite{chen-EtAl:2015:EMNLP2} &94.8&95.6& 96.5*&97.4*\\
		This work &95.5&96.5&-&-\\
		\hline
	\end{tabular}
	\caption{Comparison with previous state-of-the-art models. Results with * used external dictionary or corpus.}
\end{table*}
\subsection{Results}
\footnotetext[3]{The dictionary used in \cite{chen-EtAl:2015:ACL-IJCNLP5,chen-EtAl:2015:EMNLP2} is neither publicly released nor specified the exact source until now. We have to re-run their code using our selected dictionary to make a fair comparison. Our dictionary has been submitted along with this submission.}
We first compare our model with the latest neural network methods as shown in Table 4. The results presented in \cite{chen-EtAl:2015:ACL-IJCNLP5,chen-EtAl:2015:EMNLP2} used an extra preprocess to filter out Chinese idioms according to an external dictionary.\footnote[4]{In detail, when a dictionary is used, a preprocess is performed before training and test, which scans original text to find out Chinese idioms included in the dictionary and replace them with a unique token.} Table 4 lists the results ($\text{F}_1$-scores) with different dictionaries, which show that our models perform better when under the same settings.
\footnotetext[5]{To make comparisons fair, we re-run their code (https://github.com/dalstonChen) without using any Chinese idiom dictionary.}

Table 5 gives comparisons among previous neural network models. In the first block of Table 5, the character embedding matrix $\mathbf{M}$ is randomly initialized. The results show that our proposed novel model outperforms previous neural network methods.

Previous works have found that the performance can be improved by pre-training the character embeddings on large unlabeled data. Therefore, we use word2vec \cite{mikolov2013efficient} toolkit\footnote[6]{http://code.google.com/p/word2vec/} to pre-train the character embeddings on the Chinese Wikipedia corpus and use them for initialization. Table 5 also shows the results with additional pre-trained character embeddings. Again, our model achieves better performance than previous neural network models do.		

Table 6 compares our models with previous state-of-the-art systems. Recent systems such as \cite{zhang-EtAl:2013:EMNLP1}, \cite{chen-EtAl:2015:EMNLP2} and \cite{chen-EtAl:2015:ACL-IJCNLP5} rely on both extensive feature engineering and external corpora to boost performance. Such systems are not directly comparable with our models. In the closed-set setting, our models can achieve state-of-the-art performance on PKU dataset but a competitive result on MSR dataset, which can attribute to too strict maximum word length setting for consistence as it is well known that MSR corpus has a much longer average word length \cite{zhao2010unified}.

Table 7 demonstrates the results on MSR corpus with different maximum decoding word lengths, in which both $\text{F}_1$ scores and training time are given. The results show that the segmentation performance can indeed further be improved by allowing longer words during decoding, though longer training time are also needed. As 6-character words are allowed, $\text{F}_1$ score on MSR can be furthermore improved $0.3\%$.
\begin{table}[t]
	\centering		
	\begin{tabular}{|c|c|c|}
		\hline
		Max. word length & $\text{F}_1$ score & Time (Days)\\
		\hline
		4 & 96.5 & 4 \\
		5 & 96.7 & 5 \\
		6 & 96.8 & 6 \\
		\hline
	\end{tabular}
	\caption{Results on MSR dataset with different maximum decoding word length settings.}
\end{table}

For the running cost, we roughly report the current computation consuming on PKU dataset.\footnote[7]{Our code is released at https://github.com/jcyk/CWS.} It takes about two days to finish 50 training epochs (for results in Figure 6 and the last row of Table 6) only with two cores of an Intel i7-5960X CPU. The requirement for RAM during training is less than 800MB. The trained model can be saved within 4MB on the hard disk.
\section{Related Work}
\paragraph{Neural Network Models.} Most modern CWS methods followed \cite{xue2003chinese} treated CWS as a sequence labeling problems \cite{zhao2006effective}. Recently, researchers have tended to explore neural network based approaches \cite{collobert2011natural} to reduce efforts of feature engineering \cite{zheng-chen-xu:2013:EMNLP,qi2014deep,chen-EtAl:2015:ACL-IJCNLP5,chen-EtAl:2015:EMNLP2}. They modeled CWS as tagging problem as well, scoring tags on individual characters. In those models, tag scores are decided by context information within local windows and the sentence-level score is obtained via \textit{context-independent} tag transitions. \newcite{pei-ge-chang:2014:P14-1} introduced the tag embedding as input to capture the combinations of context and tag history. However, in previous works, only the tag of previous \textit{one} character was taken into consideration though theoretically the complete history of actions taken by the segmenter should be considered.

\paragraph{Alternatives to Sequence Labeling.} Besides sequence labeling schemes, \newcite{zhang-clark:2007:ACLMain} proposed a word-based perceptron method. \newcite{zhang2012word} used a linear-time incremental model which can also benefits from various kinds of features including word-based features. But both of them rely heavily on massive handcrafted features. Contemporary to this work, some neural models \cite{mszhang:2016,liu2016exploring} also leverage word-level information. Specifically, \newcite{liu2016exploring} use a semi-CRF taking segment-level embeddings as input and \newcite{mszhang:2016} use a transition-based framework.

Another notable exception is \cite{ma-hinrichs:2015:ACL-IJCNLP}, which is also an embedding-based model, but models CWS as configuration-action matching. However, again, this method only uses the context information within limited sized windows.

\paragraph{Other Techniques.} The proposed model might furthermore benefit from some techniques in recent state-of-the-art systems, such as semi-supervised learning \cite{zhao2008unsupervised,zhao2008exploiting,sun2011enhancing,zhao2011integrating,zeng-EtAl:2013:ACL2013,zhang-EtAl:2013:EMNLP1}, incorporating global information \cite{zhao2007incorporating,zszhang:2016}, and joint models \cite{qian2012joint,li2012unified}.
\section{Conclusion}
This paper presents a novel neural framework for the task of Chinese word segmentation, which contains three main components: (1) a factory to produce word representation when given its governed characters; (2) a sentence-level likelihood evaluation system for segmented sentence; (3) an efficient and effective algorithm to find the best segmentation. 

The proposed framework makes a latest attempt to formalize word segmentation as a direct structured learning procedure in terms of the recent distributed representation framework.

Though our system outputs results that are better than the latest neural network segmenters but comparable to all previous state-of-the-art systems, the framework remains a great of potential that can be further investigated and improved in the future.
\balance
\bibliography{acl2016}
\bibliographystyle{acl2016}
\end{document}